\documentclass[sigconf]{acmart}
\AtBeginDocument{%
  }

\usepackage{amsmath}
\usepackage{amsthm}
\usepackage{bm}
\usepackage{threeparttable}
\usepackage{multirow}
\usepackage{makecell}
\usepackage{subcaption}
\usepackage{transparent}
\usepackage[dvipsnames]{xcolor}
\allowdisplaybreaks

\newcommand{\faint}[1]{\transparent{0.65}\textcolor{gray}{#1}}
\copyrightyear{2026}
\acmYear{2026}
\setcopyright{cc}
\setcctype{by}
\acmConference[MM '26]{Proceedings of the 34th ACM International Conference on Multimedia}{November 10--14, 2026}{Rio de Janeiro, Brazil}
\acmBooktitle{Proceedings of the 34th ACM International Conference on Multimedia (MM '26), November 10--14, 2026, Rio de Janeiro, Brazil}
\acmDOI{10.1145/3767308.3834944}
\acmISBN{979-8-4007-2213-4/2026/11}

\begin{document}

\title{Ge²mS-T: Multi-Dimensional Grouping for Ultra-High Energy Efficiency in Spiking Transformer}

\author{Zecheng Hao}
\orcid{0000-0001-9074-2857}
\affiliation{
  \institution{Peking University}
  \department{School of Computer Science, State Key Laboratory for Multimedia Information Processing}
  \city{Beijing}
  \country{China}
}
\email{haozecheng@pku.edu.cn}

\author{Shenghao Xie}
\orcid{0009-0003-1024-1166}
\affiliation{
  \institution{Peking University}
  \department{School of Computer Science, State Key Laboratory for Multimedia Information Processing}
  \city{Beijing}
  \country{China}
}
\email{shenghaoxie@stu.pku.edu.cn}

\author{Kang Chen}
\orcid{0009-0001-2161-0364}
\affiliation{
  \institution{Peking University}
  \department{School of Computer Science, State Key Laboratory for Multimedia Information Processing}
  \city{Beijing}
  \country{China}
}
\email{mrchenkang@stu.pku.edu.cn}

\author{Wenxuan Liu}
\orcid{0000-0002-4417-6628}
\authornote{Corresponding author.}
\affiliation{
  \institution{Peking University}
  \department{School of Computer Science, State Key Laboratory for Multimedia Information Processing}
  \city{Beijing}
  \country{China}
}
\email{lwxfight@126.com}

\author{Zhaofei Yu}
\orcid{0000-0003-0683-6936}
\affiliation{
  \institution{Peking University}
  \department{School of Computer Science, Beijing Key Laboratory of Brain-inspired Spiking Large Models}
  \city{Beijing}
  \country{China}
}
\email{yuzf12@pku.edu.cn}

\author{Tiejun Huang}
\orcid{0000-0002-4234-6099}
\affiliation{
  \institution{Peking University}
  \department{School of Computer Science, State Key Laboratory for Multimedia Information Processing}
  \city{Beijing}
  \country{China}
}
\email{tjhuang@pku.edu.cn}

\renewcommand{\shortauthors}{Zecheng Hao et al.}

\begin{abstract}
Spiking Neural Networks (SNNs) offer superior energy efficiency over Artificial Neural Networks (ANNs). However, they encounter significant deficiencies in training and inference metrics when applied to Spiking Vision Transformers (S-ViTs). Existing paradigms including ANN-SNN Conversion and Spatial-Temporal Backpropagation (STBP) suffer from inherent limitations, precluding concurrent optimization of memory, accuracy and energy consumption. To address these issues, we propose Ge²mS-T, a novel architecture implementing grouped computation across temporal, spatial and network structure dimensions. Specifically, we introduce the Grouped-Exponential-Coding-based IF (ExpG-IF) model, enabling lossless conversion with constant training overhead and precise regulation for spike patterns. Additionally, we develop Group-wise Spiking Self-Attention (GW-SSA) to reduce computational complexity via multi-scale token grouping and multiplication-free operations within a hybrid attention-convolution framework. Experiments confirm that our method can achieve superior performance with ultra-high energy efficiency on challenging benchmarks. To our best knowledge, this is the first work to systematically establish multi-dimensional grouped computation for  resolving the triad of memory overhead, learning capability and energy budget in S-ViTs. Code is available at \url{https://github.com/hzc1208/Ge2mST}.
\end{abstract}




\maketitle
\begin{figure}[h]
  \includegraphics[width=0.96\columnwidth]{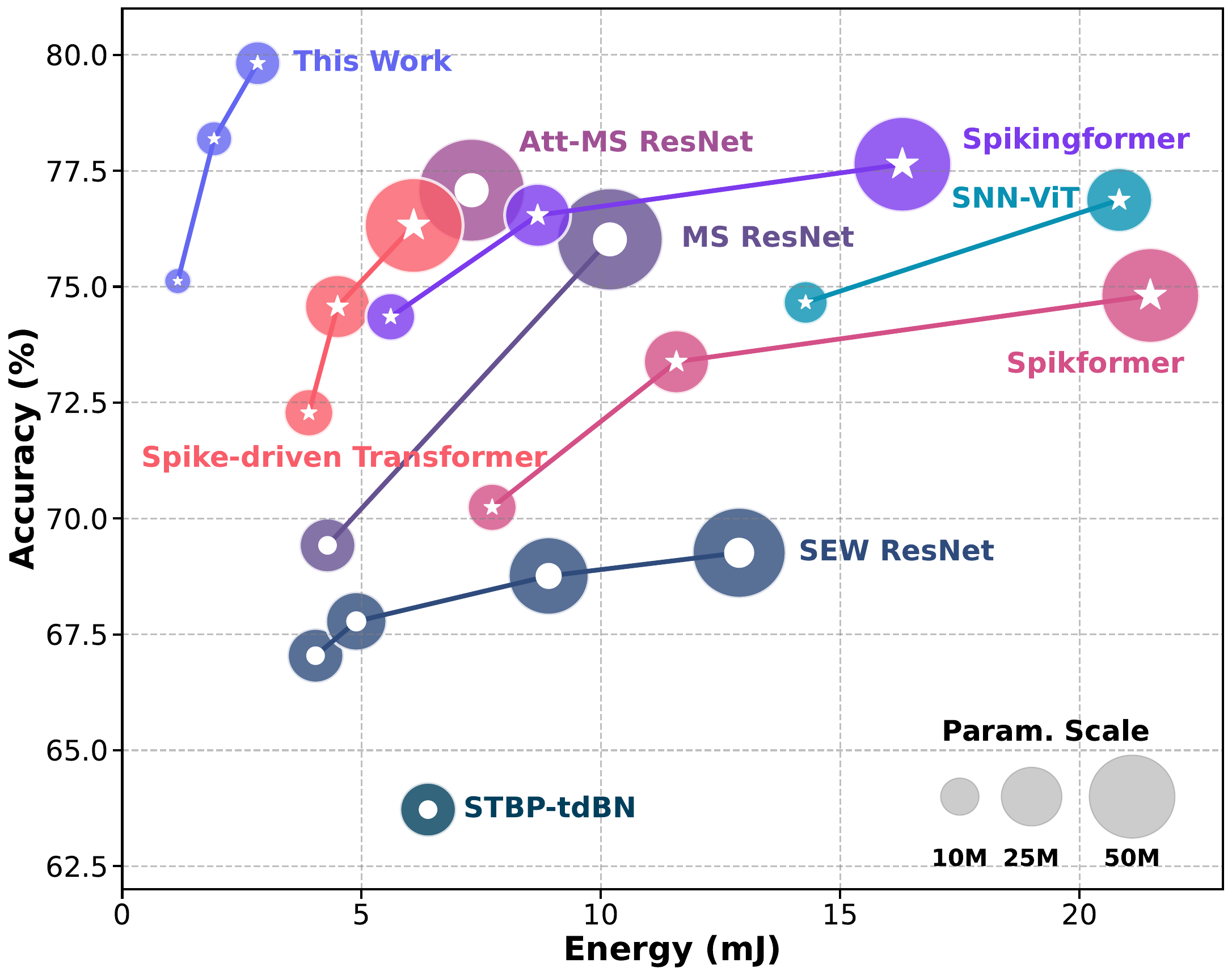}
      \caption{Comparison of network parameter count, inference accuracy and energy consumption between prior state-of-the-art (SoTA) S-ViT \& S-CNN approaches and this work on the ImageNet-1k dataset. Smaller circles denote fewer network parameters and upper-left positions indicate higher energy efficiency.}
  \label{fig01}
\end{figure}
\section{Introduction}
\begin{table*}[t]
    \caption{Performance comparison of SNN learning paradigms on S-ViT with respect to training and inference metrics.}
    \renewcommand\arraystretch{1.0}
	\centering
    \resizebox{0.85\linewidth}{!}{
    	\begin{tabular}{c|c|cc|cccc} \Xhline{0.5pt}
            \textbf{Learning Paradigm} & \textbf{Model} & \textbf{Train. Load} & \textbf{Train. Mem.} & \textbf{Inf. Latency} & \textbf{Inf. Acc.} & \textbf{SSA SOPs} & \textbf{SFFN SOPs} \\ \hline
            \faint{ANN-SNN Conversion} & \faint{IF} & \faint{$\mathcal{O}(1)$} & \faint{$\mathcal{O}(1)$} & \faint{High} & \faint{High}  & \faint{$\geq \! \mathcal{O}(T^2N^2C)$} & \faint{$\geq \! \mathcal{O}(TNC^2)$} \\
            \faint{STBP Training} & \faint{IF/LIF} & \faint{$\mathcal{O}(T)$} & \faint{$\mathcal{O}(T)$} & \faint{Low} & \faint{Low} & \faint{$\mathcal{O}(TN^2C)$} & \faint{$\mathcal{O}(TNC^2)$} \\
            \textbf{This Work} & \textbf{ExpG-IF} & $\bm{O(1)}$ & $\bm{O(1)}$ & \textbf{Low} & \textbf{High} & $\bm{O\left(\frac{TN^2C}{|G_T| |G_S|}\right)}$ & $\bm{O\left(\frac{TNC^2}{|G_T|}\right)}$ \\
            \Xhline{0.5pt}
    	\end{tabular}
    }
	\label{tab01}
\end{table*}
Spiking Neural Networks (SNNs), as the third-generation neural networks inspired by the brain structure \cite{maas1997networks}, have emerged in recent years as a novel foundational learning architecture beyond traditional Artificial Neural Networks (ANNs). Since each spiking neuron emits spikes to the post-synaptic layer only when its accumulated membrane potential after charging exceeds the firing threshold, this inherent event-driven property endows SNNs with exceptionally high sparsity at the activation level. When deployed on neuromorphic hardware \cite{ma2017darwin, davies2018loihi, debole2019truenorth, pei2019towards}, SNNs have demonstrated energy efficiency substantially superior to that of ANNs during the model inference phase.

Nevertheless, achieving high-performance training and inference for SNNs still confronts multiple challenges, primarily manifested across three dimensions: training memory, inference accuracy and energy consumption. The two mainstream learning paradigms for SNNs, namely ANN-SNN conversion \cite{cao2015spiking, li2021free, bu2022optimal, hao2025conversion} and STBP-based Training \cite{wu2018STBP, xiao2022OTTT, xu2023constructing, ding2025rethinking}, are both incapable of effectively addressing the aforementioned issues simultaneously. Consequently, exploring viable energy-efficient solutions for Spiking Vision Transformer (S-ViT), which integrate the two challenging architectures of SNNs and Transformer, has become the crowning endeavor in this field.

Prior related studies have investigated and validated the integration of conventional learning algorithms with S-ViTs. As shown in Tab.\ref{tab01}, conversion-based variants \cite{wang2023masked, huang2024towards} require only constant-order memory overhead while theoretically preserving inference accuracy upper-bounds comparable to Artificial Vision Transformer (A-ViT) \cite{liu2021swin, han2024agent}. However, error accumulation during the conversion process necessitates substantial inference time-steps to gradually recover the original network performance, and the inherent modules involving floating-point multiplications in A-ViTs render the converted S-ViTs non-native in nature, collectively posing challenges to fully realizing the energy efficiency advantages of SNNs. In contrast, STBP-based S-ViTs possess the advantage of native inference \cite{zhou2023spikformer, yao2023spike}, but the non-decouplable nature of their spatiotemporal gradients causes memory consumption to increase linearly with the number of training time-steps, while the limited capacity of vanilla spiking models to extract temporal information and the inherent backpropagation errors of approximate surrogate gradients compromise the inference accuracy of S-ViTs. Furthermore, since S-ViTs extend the temporal dimension beyond A-ViTs, the Synaptic Operations (SOPs) of their Spiking Self-Attention (SSA) and Spiking Feed-Forward Network (SFFN) during the inference phase experience substantial growth, rendering energy consumption control for S-ViTs a more critical issue compared to Spiking Convolutional Neural Network (S-CNN).

In this work, we establish grouped computation for spike sequences across temporal, spatial and network structure dimensions, proposing the architecture named \textbf{Ge²mS-T}, which aims to achieve \textbf{m}ulti-dimensional \textbf{g}rouping for ultra-high \textbf{e}nergy \textbf{e}fficiency in \textbf{S}piking \textbf{T}ransformer, simultaneously resolving the three aforementioned dilemmas hindering energy-efficient learning for S-ViTs. In the temporal dimension, inspired by non-uniform exponential quantization \cite{li2020additive}, we propose the Grouped-Exponential-Coding-based IF (ExpG-IF) model. Leveraging the philosophy of lossless conversion learning, it not only maintains constant-order training memory consumption but also achieves implicit yet precise regulation for spike firing patterns, ensuring that neurons emit spikes merely on subsets composed of specific time-step indices. In the spatial dimension, we introduce Group-wise Spiking Self-Attention (GW-SSA), which establishes a multi-scale grouping strategy for spiking tokens, effectively mitigating significant risks concerning training memory and inference energy consumption arising from the influx of substantial token quantities. Concurrently, similar with the design in A-ViTs, GW-SSA incorporates a multi-branch structure for attention and convolutional computations. Combined with the Spiking Convolutional Feed-Forward Network (Conv-SFFN), this enables Ge²mS-T to comprehensively leverage the representational advantages of both traditional S-ViT and S-CNN, while possessing a performance lower-bound comparable to that of S-CNN. It is noteworthy that the triple optimizations across the aforementioned dimensions are not mutually exclusive, their synergistic effect endows the Ge²mS-T architecture with robust competitiveness in terms of performance. Our contributions are summarized as follows:
\begin{itemize}
    \item We systematically analyze the inherent deficiencies of vanilla SSA based on ANN-SNN Conversion and STBP Training regarding relevant energy efficiency metrics, thereby substantiating the necessity and significant value of introducing the Ge²mS-T architecture.
    \item We theoretically demonstrate that the ExpG-IF model possesses the capability for lossless conversion and precise control of spike emission, while its inference computational overhead does not exceed that of the vanilla IF model.
    \item The proposed GW-SSA possesses the dual capability of capturing both global and window attention. Furthermore, its associated operations are multiplication-free and support native SNN inference, wherein combining its grouping mechanism with that of ExpG-IF achieves dual savings in inference energy consumption.
    \item Experiments validate the significant performance advantages of the Ge²mS-T architecture empowered by multi-dimensional grouping techniques. For instance, we achieve an inference accuracy of 79.82\% on ImageNet-1K, with a backbone comprising fewer than 15M parameters while consuming less than 3mJ of energy.
\end{itemize}

\section{Related Works}
\textbf{Learning algorithms for SNNs.} Currently, mainstream learning algorithms primarily encompass two categories: ANN-SNN Conversion \cite{cao2015spiking} and STBP-based Training \cite{wu2018STBP}. The former leverages the mathematical equivalence between activation functions and spiking neuron firing mechanisms, employing either training-free or fine-tuning conversion schemes based on pre-trained ANN backbones. Among these, the Quantization-Clip-Floor-Shift (QCFS) framework \cite{bu2022optimal} introduces partial constraints related to IF model computation and firing during the ANN pre-training phase, significantly enhancing the inference performance of converted QCFS-SNNs under low-latency conditions. Meanwhile, training-free conversion schemes derived from various post-training calibration methods provide feasibility for the flexible deployment of converted SNN models \cite{li2021free, hao2023bridging, wu2024ftbc}. Parallel Conversion \cite{hao2025conversion}, as an advanced variant integrating conversion learning with parallel spiking computation, achieves SNN inference with ultra-low latency and ultra-high speed.

In contrast, STBP Training represents a native SNN learning paradigm based on surrogate gradient propagation across both spatial and temporal dimensions, ensuring no performance gap between training and inference phases. However, the inherent mismatch problem of surrogate gradients \cite{guo2022recdis} and the memory consumption that grows linearly with training time-steps constitute core obstacles to the further development of STBP Training. To address these issues, STBP variants inspired by online learning effectively achieve decoupling of spatiotemporal gradients, maintaining memory consumption at a constant order of magnitude \cite{xiao2022OTTT, Meng2023SLTT}. Furthermore, ANN-SNN distillation \cite{xu2023constructing} and SNN self-distillation \cite{ding2025rethinking} frameworks ingeniously leverage implicit processes to enhance the learning performance of SNNs based on STBP Training.

\textbf{Spiking neural models.} Owing to their straightforward neural computation processes and compatibility with deep learning frameworks, LIF and IF models \cite{gerstner2002spiking} have consistently served as the cornerstone for diverse learning algorithms within the SNN community. Building upon the LIF model, on the one hand, researchers have extended its dynamical processes to propose a series of advanced spiking models with enhanced representation or memory capabilities, such as PLIF \cite{fang2020incorporating}, GLIF \cite{yao2022GLIF} and CLIF \cite{huang2024clif}; on the other hand, they have upgraded its computational mechanisms, where models like LM-HT \cite{hao2024lmht} and PSN \cite{fang2023PSN} break through the traditional unidirectional and serial computation paradigms along the temporal dimension, while multi-threshold models \cite{guo2024ternary, luo2024integer} significantly enhance the information transmission density within individual time-step and effectively optimize training memory overhead.

\textbf{Spiking Vision Transformers (S-ViT).} Currently, researchers have independently explored the feasibility of training S-ViTs along ANN-SNN conversion and STBP-based training. Regarding the former paradigm, MST \cite{wang2023masked} explores a conversion scheme based on QCFS-Transformer, whereas ECMT \cite{huang2024towards} fits pre-trained ANN backbones through multi-threshold models integrated with diverse Expectation Compensation modules. However, in general, S-ViTs derived from vanilla conversion frameworks typically suffer from severe error accumulation, necessitating substantial time-steps during the inference phase; meanwhile, training-free fitting schemes not only require manual design of specific modules, but also face compatibility challenges in practical hardware deployment due to their intricate computational procedures.

In contrast, S-ViTs obtained via STBP Training can more efficiently support inference and deployment, yet their training overhead faces the challenge of being further amplified within Transformer architectures. Spikformer \cite{zhou2023spikformer} pioneeringly introduces a network architecture with completely decoupled SSA layers and convolutional layers, while eliminating the sequential dependency in attention matrix computations, thereby mitigating the risk of memory overflow caused by the exponential growth of token quantities. Subsequently, Spike-driven Transformer (SDT) \cite{yao2023spike} explores the feasibility of linear-complexity attention blocks in SNN training, S-Resformer \cite{shi2024spikingresformer} leverages a hybrid architecture integrating SSA and Spiking Convolution (SConv) to extract richer local features for enhanced network performance, while the similarity calculation mechanism for spike sequences has also been further optimized \cite{xiao2025rethinking, guo2025spiking}. To further improve the training efficiency and reduce the energy consumption of S-ViTs, QSD-T \cite{qiu2025quantized} integrates distillation techniques to achieve stable quantization training for S-ViTs, E-Spikeformer \cite{yao2025scaling} employs Spike Firing Approximation (SFA) to establish precise alignment between the training and inference phases of S-ViTs. Inspired by binary encoding, SpikePack \cite{shen2025spikepack} realizes efficient compression and transmission for spike sequences.

\section{Preliminaries}
\textbf{LIF \& IF models} are serial computational spiking neurons characterized by the Markov property. For an inference period composed of $T$ time-steps, $\forall t\in [1,T]$, the model charges the resetted membrane potential $\mathbf{v}_{t-1}^l$ according to the input current $\mathbf{I}_t^l$. When the accumulated membrane potential $\mathbf{m}_t^l$ after charging has reached the firing threshold $\theta^l$, the model will emit spikes to the post-synaptic layer (\textit{i.e.} $\mathbf{s}_t^l=1$) and the entire firing process can be equivalently represented by the Heaviside function $\mathcal{H}(\cdot)$. Here, $\mu^l$ and $\mathbf{W}^l$ denote the membrane leaky constant and synaptic weight, respectively.
\begin{align}
    \mathbf{m}_t^l &= \mu^l \mathbf{v}_{t-1}^l + \mathbf{I}_t^l = \mu^l(\mathbf{m}_{t-1}^l \!-\! \mathbf{s}_{t-1}^l\theta^l) + \mathbf{W}^l\mathbf{s}_t^{l-1}\theta^{l-1}, \nonumber \\
    \mathbf{s}_t^l &= \mathcal{H}(\mathbf{m}_t^l-\theta^l)
    = \left\{
        \begin{aligned}
        &1,\ \mathbf{m}_t^l \geq \theta^l \\
        &0,\ \text{otherwise}
        \end{aligned}\right..
    \label{eq01}
\end{align}
\textbf{ANN-SNN Conversion} is established upon the equivalent mapping relationship possessed by the IF model (\textit{i.e.} $\mu^l=1$) based on the soft-reset mechanism (\textit{i.e.} $\mathbf{v}_t^l=\mathbf{m}_t^l-\mathbf{s}_t^l\theta^l$). Consequently, based on the aforementioned conditions and Eq.\eqref{eq01}, we can derive $\mathbf{v}_t^l=\mathbf{v}_{t-1}^l+\mathbf{W}^l\mathbf{s}_t^{l-1}\theta^{l-1}-\mathbf{s}_t^l\theta^l$. Subsequently, accumulating and averaging along the temporal dimension yields the following equation:
\begin{align}
    \mathbf{F}_\text{Avg}^l &= \frac{1}{T} \sum_{t=1}^T \mathbf{s}_t^l\theta^l = \frac{1}{T} \left(\sum_{t=1}^T \mathbf{W}^l\mathbf{s}_t^{l-1}\theta^{l-1} \!+\! \mathbf{v}_{t-1}^l \!-\! \mathbf{v}_t^l \right) \nonumber \\
    &= \mathbf{W}^l \mathbf{F}_\text{Avg}^{l-1} - \frac{\mathbf{v}_T^l \!-\! \mathbf{v}_0^l}{T} = \mathbf{I}_\text{Avg}^l - \frac{\mathbf{v}_T^l \!-\! \mathbf{v}_0^l}{T}.
    \label{eq02}
\end{align}
One can note that the average firing rates of adjacent layers $\mathbf{F}_\text{Avg}^l, \mathbf{F}_\text{Avg}^{l-1}$ adhere to an approximately linear transformation relationship. In particular, for the ideal case where $T\rightarrow +\infty$, we are able to obtain a pre-trained ANN backbone with $\mathcal{O}(1)$ computational overhead during the training phase and convert its activation layers into the IF model layer-by-layer without conversion errors during the inference phase.

\textbf{STBP Training} involves performing gradient computation for the entire dynamical process in Eq.\eqref{eq01} by formulating it as Recurrent Neural Networks (RNNs). Throughout the entire backpropagation chain, the primary challenge lies in addressing the non-differentiability of $\mathbf{s}_t^l=\mathcal{H}(\mathbf{m}_t^l-\theta^l)$, current studies primarily employ the surrogate gradient approach (\textit{e.g.} $\frac{\partial \mathbf{s}_t^l}{\partial \mathbf{m}_t^l}=\max(1-|\mathbf{m}_t^l-\theta^l|,0)$) to facilitate approximate computation.

\textbf{Conversion-based \& STBP-based SSAs.} Regarding the core process of attention computation, we define $\mathbf{Q}^l,\mathbf{K}^l,\mathbf{V}^l \in\mathbb{R}^{T\times B\times N\times C}$, where $B,N,C$ respectively denote the batch size, number of spiking tokens and feature dimension. $f(\cdot)$ represents the activation module applied to the attention matrix.
\begin{align}
    & \textbf{Attn}^l\big|_\text{A2S} \! = \! f \left(\mathbf{Q}_\text{Avg}^l {\mathbf{K}_\text{Avg}^l}^{\text{T}} \right) \mathbf{V}_\text{Avg}^l \!=\! \left\{ \! \frac{1}{T^2} f \left( \mathbf{Q}_t^l\sum_{i=1}^T {\mathbf{K}_i^l}^{\text{T}} \! \right) \sum_{j=1}^T\mathbf{V}_j^l \right\}, \nonumber \\
    & \textbf{Attn}^l\big|_\text{STBP} = \left\{ f \left(\mathbf{Q}_1^l {\mathbf{K}_1^l}^{\text{T}} \right) \mathbf{V}_1^l,\ \cdots \ , f \left(\mathbf{Q}_T^l {\mathbf{K}_T^l}^{\text{T}} \right) \mathbf{V}_T^l \right\}.
    \label{eq03}
\end{align}
In the context of ANN-SNN Conversion, the pre-training phase essentially simulates the firing rates at corresponding positions as $\mathbf{Q}_\text{Avg}^l,\mathbf{K}_\text{Avg}^l,\mathbf{V}_\text{Avg}^l \in\mathbb{R}^{1\times B\times N\times C}$, yet this necessitates a computational overhead of $\mathcal{O}(T^2N^2C)$ during the inference phase to achieve alignment with the predicted firing rates, as illustrated in Eq.\eqref{eq03}. Furthermore, given that practical conversion errors cannot be fully eliminated within $T$ time-steps, effectively recovering the accuracy of the converted S-ViT further exacerbates the rise in inference energy consumption. Conversely, regarding STBP Training, attention computation is performed exclusively between feature maps within the same time-step (\textit{i.e.} $\forall t\in[1,T],f \left(\mathbf{Q}_t^l {\mathbf{K}_t^l}^{\text{T}} \right) \mathbf{V}_t^l$), while this avoids attention computation of $\mathcal{O}(T^2)$ complexity, the characteristics of STBP coupled with the absence of attention interaction across time-steps render STBP-based SSA inferior in terms of both training overhead and inference accuracy.

\section{Methods}
\begin{figure*}[t]
  \includegraphics[width=0.85\linewidth]{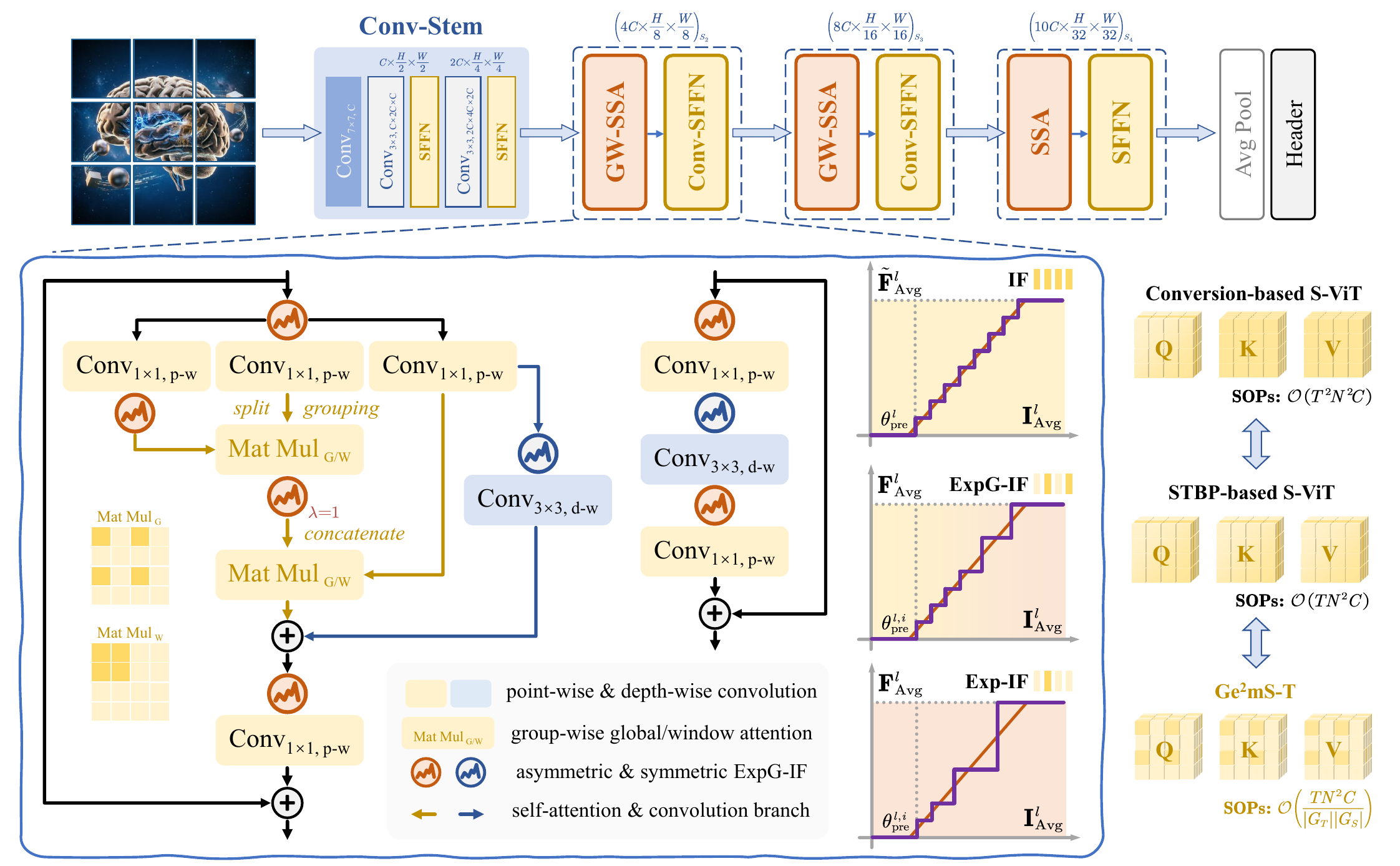}
      \caption{Overall architecture of Ge²mS-T, comprising a series of stage-oriented blocks constructed from SSA, SConv and SFFN. Compared with Conversion-based and STBP-based SSAs, the energy efficiency of GW-SSA is substantially enhanced across multiple dimensions.}
  \label{fig02}
\end{figure*}
\subsection{ExpG-IF: Temporal Dimension Grouping for Spike Sequences}
For vanilla spiking models under the ANN-SNN Conversion or STBP Training frameworks, researchers currently still struggle to identify feasible solutions capable of effectively reducing the number of spike emissions while preserving the overall learning performance of the network. Herein, we innovatively establish an equivalent mapping relationship between such sequences and non-uniform quantization activation functions within the exponential coding space oriented toward spike sequences, which not only enables implicit yet precise regulation of the spike firing count, but also supports lossless conversion under $\mathcal{O}(1)$ training overhead, thereby allowing the SNN to fully inherit the learning capabilities of the pre-trained ANN during the inference phase.

More precisely, conventional conversion learning oriented toward the vanilla IF model typically simulates and predicts the spike firing rate based on the following function:
\begin{align}
    \mathbf{\tilde{F}}_\text{Avg}^l = \frac{\theta^l}{T} \text{clip} \left( \left\lfloor \frac{\mathbf{I}_\text{Avg}^l T+\mathbf{v}_0^l}{\theta^l} \right\rfloor,0,T \right) \ \Rightarrow \ \mathbf{s}^l= \left\{ \mathbf{s}_1^l,\ \cdots\ , \mathbf{s}_T^l \right\}.
    \label{eq04}
\end{align}
In other words, the simulated firing rate $\mathbf{\tilde{F}}_\text{Avg}^l$ during the training phase is estimated directly via uniform quantization based on $\mathbf{I}_\text{Avg}^l$ \cite{bu2022optimal}. In comparison, within the ExpG-IF model, we aim to simulate the firing rate based on $\mathbf{I}_\text{Avg}^l$ in the form of non-uniform quantization and subsequently transmit the spike sequence in the form of exponential coding during the SNN inference phase:
\begin{align}
    & \mathbf{\tilde{F}}_\text{Avg}^l = \frac{\lambda^l (1 \!+\! \cos\!\frac{\pi\mathbf{e}_\text{t}}{\mathbf{e}_\text{T}})}{2} \text{clip}\! \left( \! \left\lfloor \frac{\mathbf{I}_\text{Avg}^l T \!+\! \mathbf{v}_0^l}{\lambda^l} \right\rceil \!,\! 0,1 \! \right) \!+\! \frac{(1 \!-\! \cos\!\frac{\pi\mathbf{e}_\text{t}}{\mathbf{e}_\text{T}})}{2} \mathbf{F}_\text{Avg}^l, \nonumber 
    \\
    &\mathbf{F}_\text{Avg}^l \!=\! \frac{\lambda^l}{\mathbf{s}_{\max}^l} \text{clip}\! \left( \! \left\lfloor \frac{\mathbf{I}_\text{Avg}^l T \!+\! \mathbf{v}_0^l}{\lambda^l} \right\rceil_{\bm{\theta}^l} \!,\! 0,\mathbf{s}_{\max}^l \! \right) \!\! \Leftrightarrow\! \mathbf{s}^l \!=\! \left\{ \alpha^0 \mathbf{s}_1^l,\cdots, \alpha^{T-1} \mathbf{s}_T^l \right\}, \nonumber 
    \\
    &\bm{\theta}^l \!=\! \left\{\! \sum_{i=1}^n \! x_i \Bigg| x_i \!\in\! G_T^{(i)}, \bigcup_{i=1}^n \! G_T^{(i)} \!=\! \left\{ 0,\! \alpha^0 \!,\!\cdots\!,\! \alpha^{T-1} \right\} \!\land\! \bigcap_{i=1}^n \! G_T^{(i)} \!=\! \left\{ 0 \right\} \! \right\} \!.
    \label{eq05}
\end{align}
Here $\mathbf{s}_{\max}^l=\max(\bm{\theta}^l)$ and $\lfloor\cdot\rceil_{\bm{\theta}^l}$ encompasses $|\bm{\theta}^l|$ quantization levels whose specific values correspond sequentially to the elements within $\bm{\theta}^l$. As observed from Eq.\eqref{eq05}, the $T$ bases $\left\{ \alpha^0, \cdots, \alpha^{T-1} \right\}$ within the current exponential coding are allocated into $n$ distinct groups, enabling $\bm{\theta}^l$ to selectively cancel certain levels from the original uniform quantization. Furthermore, the effect of mapping the aforementioned operations to the SNN inference phase lies in implicitly restricting the maximum spike firing count of the ExpG-IF model within $T$ time-steps to not exceed $n$. Specially, when $n=1$, the firing rate during the training phase is modeled entirely according to an exponential scale, we designate the spiking model in this scenario as Exp-IF. Acknowledging that directly training S-ViT based on non-uniformly quantized activations presents significant challenges, we adopt a progressive training scheme grounded in mixed activation functions which transitions gradually from the high-precision ClipReLU function throughout the training process to the final state represented by $\mathbf{F}_\text{Avg}^l$. Here, $\mathbf{e}_\text{t},\mathbf{e}_\text{T}$ denote the current and total number of training epochs, respectively. 

The conversion schemes concerning the firing threshold differ between ExpG-IF and vanilla IF model. Specifically, we can further refine $\theta^l$ into $\theta^l_\text{pre},\theta^l_\text{post}$, where $\mathbf{s}_t^l=\mathcal{H}(\mathbf{m}_t^l-\theta^l_\text{pre}),\mathbf{I}_t^{l+1}=\mathbf{W}^{l+1}\mathbf{s}_t^l\theta^l_\text{post}$. For conversion learning based on vanilla IF, $\theta^l_\text{pre},\theta^l_\text{post}$ can be uniformly set as the learnable scaling factor $\lambda^l$. In contrast, ExpG-IF incorporates multiple post-firing thresholds where $\theta^{l,i}_\text{post}\in\lambda^l\bm{\theta}^l$, and the pre-firing threshold is configured correspondingly based on the quantization rule $\lfloor\cdot\rceil_{\bm{\theta}^l}$ during the training phase such that $\theta_\text{pre}^{l,i}\in \lambda^l(\bm{\theta}^l_{:|\bm{\theta}^l|}+\bm{\theta}^l_{1:})/2$.
In addition, it is noteworthy that this regulation relationship is confined to the exponential coding environment: for instance, if quantization level $k$ is canceled within the coding environment of Eq.\eqref{eq04}, the originally corresponding $\mathbf{I}_\text{Avg}^l$ interval will be reassigned to $k-1$ or $k+1$, yet it cannot be theoretically guaranteed that the spike firing count after reassignment will decrease.

In terms of inference computational overhead, ExpG-IF enables direct querying of the corresponding spike pattern based on $\mathbf{I}_\text{Avg}^l$, which is completely lossless compared to the training phase. Given $|\bm{\theta}^l|\leq 2^{T}$, utilizing binary search to identify the specific interval position requires merely $\mathcal{O}(T)$ comparison operations while computing $\mathbf{I}_\text{Avg}^l$ itself demands $T-1$ addition operations, thus the total computational cost does not surpass that of the vanilla IF model, which comprises $\mathcal{O}(T)$ comparison operations for firing and $2T$ addition operations for charging and resetting.

\subsection{GW-SSA: Multi-scale Spatial Dimension Grouping for Spiking Tokens}
\begin{table*}[t]
    \caption{Detailed configurations of Ge²mS-T on the ImageNet-1k dataset. Here $R_1$ denotes the expansion ratio of SConv within $\text{Conv}_\text{B}$, $R_2,R$ denote the expansion ratio of Conv-SFFN (SFFN), $r_\text{g,w},h$ represent the split ratio of global \& window attention dimensions in GW-SSA and the number of attention heads.}
    \renewcommand\arraystretch{1.08}
	\centering
    \resizebox{0.9\linewidth}{!}{
    	\begin{tabular}{cccccc} \Xhline{0.5pt}
            \textbf{Stage} & \textbf{Feature Size} & \textbf{Block Structure} & \textbf{Ge²mS-T Small} & \textbf{Ge²mS-T Base} & \textbf{Ge²mS-T Large} \\
            \hline
            Stem & 224$\times$224 & Convolution & \multicolumn{3}{c}{Conv 7$\times$7, stride 2} \\
            \hline
            \multirow{2}{*}{Stage 1} & 112$\times$112 & ExpG-IF & $\begin{bmatrix} C=24,R_1,R_2=2,4 \end{bmatrix}$ & $\begin{bmatrix} C=32,R_1,R_2=2,4 \end{bmatrix}$ &  $\begin{bmatrix} C=40,R_1,R_2=2,4 \end{bmatrix}$ \\
            & 56$\times$56 & SConv, SFFN & $\begin{bmatrix} C=48,R_1,R_2=2,4 \end{bmatrix}$ & $\begin{bmatrix} C=64,R_1,R_2=2,4 \end{bmatrix}$ & $\begin{bmatrix} C=80,R_1,R_2=2,4 \end{bmatrix}$ \\ \hline

            \multirow{3}{*}{Stage 2} & \multirow{3}{*}{28$\times$28} & ExpG-IF & \multirow{3}{*}{$\begin{bmatrix} C=96, r_\text{g,w}=0.5 \\ |G_S|=4^2, |G_T|=2 \\ h=4, R=4 \end{bmatrix} \times 2$} & \multirow{3}{*}{$\begin{bmatrix} C=128, r_\text{g,w}=0.5 \\ |G_S|=4^2, |G_T|=2 \\ h=4, R=4 \end{bmatrix} \times 2$} &  \multirow{3}{*}{$\begin{bmatrix} C=160, r_\text{g,w}=0.5 \\ |G_S|=4^2, |G_T|=2 \\ h=4, R=4 \end{bmatrix} \times 2$} \\
            & & GW-SSA & & & \\ 
            & & Conv-SFFN & & & \\ \hline

            \multirow{3}{*}{Stage 3} & \multirow{3}{*}{14$\times$14} & ExpG-IF & \multirow{3}{*}{$\begin{bmatrix} C=192, r_\text{g,w}=0.5 \\ |G_S|=2^2, |G_T|=2 \\ h=8, R=4 \end{bmatrix} \times 6$} & \multirow{3}{*}{$\begin{bmatrix} C=256, r_\text{g,w}=0.5 \\ |G_S|=2^2, |G_T|=2 \\ h=8, R=4 \end{bmatrix} \times 6$} &  \multirow{3}{*}{$\begin{bmatrix} C=320, r_\text{g,w}=0.5 \\ |G_S|=2^2, |G_T|=2 \\ h=8, R=4 \end{bmatrix} \times 6$} \\
            & & GW-SSA & & & \\
            & & Conv-SFFN & & & \\ \hline

            \multirow{2}{*}{Stage 4} & \multirow{2}{*}{7$\times$7} & ExpG-IF & \multirow{2}{*}{$\begin{bmatrix} C=240 \\ h=8, R=4 \end{bmatrix} \times 2$} & \multirow{2}{*}{$\begin{bmatrix} C=320 \\ h=8, R=4 \end{bmatrix} \times 2$} &  \multirow{2}{*}{$\begin{bmatrix} C=400 \\ h=8, R=4 \end{bmatrix} \times 2$} \\
            & & SSA, SFFN & & & \\ \hline
            
            Classifier & 1$\times$1 & Linear & \multicolumn{3}{c}{1000-FC} \\
            \Xhline{0.5pt}
    	\end{tabular}
    }
	\label{tab02}
\end{table*}
The aforementioned discussion has indicated that the computational overhead of vanilla SSA in Eq.\eqref{eq03} is no less than $\mathcal{O}(TN^2C)$, which causes resource constraints for S-ViTs based on this mechanism when processing feature maps with a large number of spiking tokens. Herein GW-SSA achieves energy-efficient training and inference for SSA by grouping tokens $\mathbf{X}^l$ along the spatial dimension and performing attention computation exclusively within groups. Considering that the direct grouping attention mechanism may compromise the performance upper-bound of S-ViTs due to the absence of interaction between inter-group tokens, we further propose a multi-scale grouping version oriented toward GW-SSA.
\begin{align}
    &\textbf{Attn}^l \!=\! \left\{ f \left(\mathbf{Q}^{l,(\text{g})} {\mathbf{K}^{l,(\text{g})}}^{\text{T}} \right)  \mathbf{V}^{l,(\text{g})}, f \left(\mathbf{Q}^{l,(\text{w})} {\mathbf{K}^{l,(\text{w})}}^{\text{T}} \right) \mathbf{V}^{l,(\text{w})} \right\}, \nonumber \\
    &\mathbf{X}^l \!=\! \left\{ \! \bigcup_{i,j=1}^{n} \!\! G_S^{(ij)} \Bigg| \mathbf{X}_{(i::n,j::n,:c)}^l \!\! \in\! G_S^{(\text{g})}, \mathbf{X}_{\left( \! \frac{(i-1)H}{n}:\frac{iH}{n},\frac{(j-1)W}{n}:\frac{jW}{n},c: \! \right)}^l \!\!\in\! G_S^{(\text{w})} \!\! \right\}.
    \label{eq06}
\end{align}
As shown in Eq.\eqref{eq06}, for $\mathbf{X}^l\in \mathbb{R}^{T\times B\times H\times W\times C}$ (where $N=H\times W$ in S-ViT with focus on the latter three dimensions), we first split along the channel dimension such that $\mathbf{X}^{l,(\text{g})}=\mathbf{X}_{:c}^l$ is oriented toward approximate global attention computation, while $\mathbf{X}^{l,(\text{w})}=\mathbf{X}_{c:}^l$ is oriented toward local window-based attention calculation. Here $\mathbf{X}^{l,(\text{g})}$ undergoes further grouping along the height and width dimensions. For instance, traversing the aforementioned two dimensions with a stride of $n$ yields $|G_S^{(\text{g})}|=n^2$ token subsets, where each subset constitutes a pooled and approximate global feature map followed by SSA computation within each subset. Conversely, the specific grouping strategy for $\mathbf{X}^{l,(\text{w})}$ differs as it directly splits the aforementioned dimensions into $n$ chunks respectively to obtain $n^2$ local windows, then executing SSA calculation within each window. Upon completion of the aforementioned $2n^2$ groups of attention computation, the results are concatenated along these three dimensions to restore the original shape such that $\textbf{Attn}^l\in \mathbb{R}^{T\times B\times H\times W\times C}$. At this point, the computational overhead of the overall process decreases from $\mathcal{O}(N^2C)$ to $\mathcal{O}(\frac{N^2C}{|G_s|})$.
\subsection{Ultra-High Energy-Efficient S-ViT Architecture via Multi-Dimensional Grouping}
In this section, we provide a detailed elaboration on the relevant components of Ge²mS-T, as illustrated in Fig.\ref{fig02} and Tab.\ref{tab02}. Here down-sampling and average pooling layers are omitted for clarity. For given input data $\mathbf{I}_x$, local features are initially extracted through a convolution-based Conv-Stem, since shallow feature maps typically possess an excessively large number of spiking tokens, rendering the computational overhead substantial even when executing GW-SSA. Consequently, we employ a two-layer SConv involving channel upsampling to replace the GW-SSA module here, which combines with the SFFN module to constitute the $\text{Conv}_\text{B}$ block.
\begin{align}
    & \text{Conv-Stem}(\mathbf{I}_x) = \text{Conv}_\text{B}^{(S_1)} \left(\text{Conv}(\mathbf{I}_x) \right), \nonumber \\
    & \text{Conv}_\text{B}(\mathbf{I}^l) = \text{SFFN}_\text{Conv} \left(\text{Conv}(\text{SN}(\text{Conv}(\text{SN}(\mathbf{I}^l)))) + \mathbf{I}^l \right).
    \label{eq07}
\end{align}
Eqs.(\ref{eq08}-\ref{eq09}) illustrate the specific structures of GW-SSA and Conv-SFFN respectively. Regarding GW-SSA, we design a dual-branch pathway based on attention and convolution. Here $\text{SN}(\cdot),\text{SN}_\text{sym}(\cdot)$ denote the vanilla ExpG-IF model and a variant version with positive-negative symmetric numerical intervals, which are applicable to positions requiring and not requiring activation functions respectively. Specifically, the attention pathway combines the temporal dimension grouping for spike patterns in Eq.\eqref{eq05} with the multi-scale grouping attention mechanism described in Eq.\eqref{eq06} to achieve multiple enhancements in energy efficiency. To circumvent the computational burden associated with the $\mathcal{O}(T^2)$ complexity of SSA akin to conversion learning, we average $\mathbf{K}^l,\mathbf{V}^l$ along the temporal dimension, this modification theoretically exerts no influence on the network architecture based on ExpG-IF. The activation function $f(\cdot)$ for the attention matrix is also replaced by $\text{SN}(\cdot)$, ensuring that the entire attention computation process fully supports multiplication-free operations. Meanwhile, the convolution pathway further extracts features based on $\mathbf{V}^l$ and is subsequently added to the attention pathway. This ensures that from the perspective of treating $\mathbf{V}^l$ as the backbone pathway, GW-SSA integrates global, windowed and local weighted information for spiking tokens, guaranteeing that its learning capability is no lower than the performance lower-bounds of vanilla SSA and S-CNN. 
\begin{align}
    & \mathbf{Q}^l, (\mathbf{K}^l, \mathbf{V}^l) = \text{SN}\! \left( \text{Conv}_\text{p-w}( \text{SN}(\mathbf{I}^l ) ) \right), \text{Conv}_\text{p-w}\left( \text{SN}(\mathbf{I}^l ) \right) \nonumber \\
    & \textbf{Attn}^l \!=\! \left\{ \text{SN}\! \left( \! \mathbf{Q}^{l,(\text{g})} {\mathbf{K}_\text{Avg}^{l,(\text{g})}}^{\text{T}} \right)\!  \mathbf{V}_\text{Avg}^{l,(\text{g})}, \text{SN}\! \left( \! \mathbf{Q}^{l,(\text{w})} {\mathbf{K}_\text{Avg}^{l,(\text{w})}}^{\text{T}} \right)\! \mathbf{V}_\text{Avg}^{l,(\text{w})} \right\}, \nonumber \\
    & \mathbf{O}^l = \text{Conv}_\text{p-w}\left( \text{SN}( \textbf{Attn}^l + \text{Conv}_\text{d-w}(\text{SN}_\text{sym}(\mathbf{V}^l)) \right) + \mathbf{I}^l. \label{eq08} \\
    &\mathbf{H}^{l+1}= \text{Conv}_\text{d-w} \left(\text{SN}_\text{sym}(\text{Conv}_\text{p-w}(\text{SN}(\mathbf{I}^{l+1} ))) \right), \nonumber \\
    &\mathbf{O}^{l+1} = \text{Conv}_\text{p-w} \left(\text{SN}(\mathbf{H}^{l+1}) \right) + \mathbf{I}^{l+1}.
    \label{eq09}
\end{align}
\begin{table*}[t]
    \caption{Performance evaluation of relevant approaches based on different learning paradigms and network architectures on the ImageNet-1k dataset. Here SOPs denotes the average number of synaptic operations per input sample, and Energy is computed following the calculation standards outlined in \cite{zhou2023spikformer}.}
    \renewcommand\arraystretch{1.0}
	\centering
    \resizebox{0.93\linewidth}{!}{
      \begin{threeparttable}
	   \begin{tabular}{cc|cccccc} \Xhline{1pt}
        \textbf{Method} & \textbf{Type} & \textbf{Architecture} & \textbf{Param.(M)} & \textbf{T} & \textbf{SOPs(G)} & \textbf{Energy(mJ)} & \textbf{Acc.(\%)} \\ \hline

        \multirow{2}{*}{Spiking ResNet \cite{hu2021residual}} & \multirow{2}{*}{ANN-SNN Conversion} & ResNet-34 & 21.79 & 350 & 65.28 & 59.30 & 71.61 \\
        & & ResNet-50 & 25.56 & 350 & 78.29 & 70.93 & 72.75 \\ \hline
        
        STBP-tdBN \cite{zheng2021going} & STBP Training & Spiking ResNet-34 & 21.79 & 6 & 6.50 & 6.39 & 63.72 \\
        \multirow{4}{*}{SEW ResNet \cite{fang2021deep}} & \multirow{4}{*}{STBP Training} & SEW ResNet-34 & 21.79 & 4 & 3.88 & 4.04 & 67.04 \\
        & & SEW ResNet-50 & 25.56 & 4 & 4.83 & 4.89 & 67.78 \\
        & & SEW ResNet-101 & 44.55 & 4 & 9.30 & 8.91 & 68.76 \\
        & & SEW ResNet-152 & 60.19 & 4 & 13.72 & 12.89 & 69.26 \\
        \multirow{2}{*}{MS ResNet \cite{hu2024residual}} & \multirow{2}{*}{STBP Training} & MS ResNet-34 & 21.79 & 6 & 4.77 & 4.29 & 69.42 \\
        & & MS ResNet-104\tnote{$\dag$} & 77.28 & 5 & 11.32 & 10.19 & 76.02 \\
        Att-MS ResNet \cite{yao2023attention} & STBP Training & MS ResNet-104\tnote{$\dag$} & 78.37 & 4 & - & 7.30 & 77.08 \\ \hline

        \multirow{3}{*}{Spikformer \cite{zhou2023spikformer}} & \multirow{3}{*}{STBP Training} & Spikformer-8-384 & 16.81 & 4 & 6.82 & 7.73 & 70.24 \\
        & & Spikformer-8-512 & 29.68 & 4 & 11.09 & 11.58 & 73.38 \\
        & & Spikformer-8-768 & 66.34 & 4 & 22.09 & 21.48 & 74.81 \\
        \multirow{3}{*}{SDT \cite{yao2023spike}} & \multirow{3}{*}{STBP Training} & SDT-8-384 & 16.81 & 4 & - & 3.90 & 72.28 \\
        & & SDT-8-512 & 29.68 & 4 & - & 4.50 & 74.57 \\
        & & SDT-8-768 & 66.34 & 4 & - & 6.09 & 76.32 \\ 
        \multirow{2}{*}{SNN-ViT \cite{wang2025svit}} & \multirow{2}{*}{STBP Training} & SViT-8-256 & 13.7 & 4 & - & 14.28 & 74.66 \\
        & & SViT-8-384 & 30.4 & 4 & - & 20.83 & 76.87 \\        
        \multirow{3}{*}{Spikingformer \cite{zhou2026spikingformer}} & \multirow{3}{*}{STBP Training} & Spikingformer-8-384 & 16.81 & 4 & - & 5.61 & 74.35 \\
        & & Spikingformer-8-512 & 29.68 & 4 & - & 8.68 & 76.54 \\
        & & Spikingformer-8-768 & 66.34 & 4 & - & 16.30 & 77.64 \\ \hline

        \multirow{3}{*}{\textbf{This Work}} & \multirow{3}{*}{\textbf{\makecell{Ge²mS-T Training}}} & Ge²mS-T Small & \textbf{5.35} & 4 & \textbf{1.29} & \textbf{1.16} & \textbf{75.12} \\
        & & Ge²mS-T Base & \textbf{9.36} & 4 & \textbf{2.14} & \textbf{1.92} & \textbf{78.19} \\
        & & Ge²mS-T Large & \textbf{14.48} & 4 & \textbf{3.15} & \textbf{2.83} & \textbf{79.82} \\
        
        \Xhline{1pt}
	    \end{tabular}
        \begin{tablenotes}
            \small
            \item[$\dag$] denotes enlarged 288$\times$288 input size for inference; default is 224$\times$224 for training and inference stages.
        \end{tablenotes}
        \end{threeparttable}
    }
	\label{tab03}
\end{table*}
Conv-SFFN is based on the architecture of depthwise separable convolution, simultaneously possessing the capacity to store representational knowledge and extract local information. Cooperating with the dual-branch structure of GW-SSA, this enables the $\text{SSA}_\text{B}$ block to achieve hybrid driving by convolution and attention layers. Within Ge²mS-T, the input current passes through a series of $\text{SSA}_\text{B}$ blocks that are divided into three stages based on different input scales. In this context, the number of spiking tokens upon entering the third stage has been compressed to a minimal amount, obviating the need for further local information extraction, so we directly employ vanilla SSA and SFFN under the same architecture followed by connecting a global pooling layer and an output header. 
\begin{align}
    & \text{Ge}^{2}\text{mS-T}(\mathbf{I}_x) = \text{Header} \left(\text{SSA}_\text{B}^{(S_2,S_3,S_4)}( \text{Conv-Stem}(\mathbf{I}_x)) \right), \nonumber \\
    & \text{SSA}_\text{B}(\mathbf{I}^l) \in \left\{ \text{SFFN}_\text{Conv} \left(\text{SSA}_\text{GW}(\mathbf{I}^l) \right), \text{SFFN}\left(\text{SSA}(\mathbf{I}^l) \right) \right\}.
    \label{eq10}
\end{align}
From the perspective of the overall architecture, Ge²mS-T incorporates the characteristics of the aforementioned three dimensions while simultaneously breaking through the inherent bottlenecks of S-ViTs across multiple performance metrics.

\section{Experiments}
In this section, we integrate relevant approaches towards S-CNN and S-ViT architectures under various learning paradigms as our comparative targets, conducting comprehensive validation of energy-efficiency-related metrics across a series of challenging benchmarks comprising both static and neuromorphic datasets, including ImageNet-1k \cite{Deng2009ImageNet}, CIFAR-10(100) \cite{Krizhevsky2009CIFAR100} and CIFAR10-DVS \cite{li2017cifar10}, with evaluation metrics encompassing SOPs, Energy, Accuracy and Firing Rate. More detailed experimental configuration is provided in the supplementary materials.

\subsection{Performance Comparison with Prior SoTA Works}
As presented in Tab.\ref{tab03}, we systematically summarize representative works based on various foundational SNN architectures, including vanilla Spiking ResNet \cite{hu2021residual}, SEW ResNet \cite{fang2021deep}, MS ResNet \cite{hu2024residual}, MS ResNet equipped with lightweight spiking attention mechanisms \cite{yao2023attention}, SEW-based SSA \cite{zhou2023spikformer}, linear-complexity SSA \cite{yao2023spike}, causal SSA along the temporal dimension \cite{wang2025svit} and MS-based SSA \cite{zhou2026spikingformer}.

Regarding S-CNN architectures, Ge²mS-T Small achieves an additional accuracy improvement of 11.40\% with merely 24.55\% of the parameters and 18.15\% of the energy consumption of Spiking ResNet-34. Compared with SEW ResNet-152, Ge²mS-T Base attains an accuracy advantage of 8.93\% while consuming merely 15.55\% parameters and 14.90\% energy, respectively. Ge²mS-T Large surpasses MS ResNet-104 without and with lightweight attention mechanisms by 3.80\% and 2.74\% in accuracy, respectively, while requiring only 18.74\%, 18.48\% of parameters and 27.77\%, 38.77\% of energy consumption.

For S-ViT architectures, Ge²mS-T Small achieves superior accuracy while requiring only 8.06\% of the parameters and 5.40\% of the energy consumption compared to Spikformer-8-768. Ge²mS-T Base exceeds SDT-8-384 by 5.91\% in accuracy while utilizing 55.68\% of parameters and 49.23\% of energy, and achieves a 3.53\% accuracy gain over SViT-8-256 while consuming merely 13.45\% of its energy budget. Ge²mS-T Large occupies merely 48.79\% (21.83\%) of parameters and 32.60\% (17.36\%) of energy relative to Spikingformer-8-512 (768), yet still attains corresponding accuracy improvements of 3.28\% (2.18\%).

\subsection{Validation on Downstream Benchmarks}
\begin{table*}[t]
    \caption{Performance evaluation on downstream benchmarks, which encompass both static and neuromorphic datasets, including CIFAR-10, CIFAR-100 and CIFAR10-DVS.}
    \renewcommand\arraystretch{1.0}
	\centering
    \resizebox{0.85\linewidth}{!}{
    	\begin{tabular}{cc|cccccc} \Xhline{1pt}
            \multirow{2}{*}{\textbf{Method}} & \multirow{2}{*}{\textbf{Type}} & \multicolumn{2}{c}{\textbf{CIFAR-10}} & \multicolumn{2}{c}{\textbf{CIFAR-100}} & \multicolumn{2}{c}{\textbf{CIFAR10-DVS}} \\ \cline{3-8}
            & & \textbf{Time-step} & \textbf{Acc.(\%)} & \textbf{Time-step} & \textbf{Acc.(\%)} & \textbf{Time-step} & \textbf{Acc.(\%)} \\ \hline
            STBP-tdBN \cite{zheng2021going} & STBP Training & 4 & 92.92 & - & - & 10 & 67.8 \\
            TET \cite{deng2022temporal} & STBP Training & 4 & 94.44 & 4 & 74.47 & 10 & 83.17 \\
            SLTT \cite{Meng2023SLTT} & Online Training & 6 & 94.44 & 6 & 74.38 & 10 & 82.20 \\
            GAC-SNN \cite{qiu2024gated} & STBP Training & 4 & 96.24 & 4 & 79.83 & - & - \\ \hline

            \multirow{2}{*}{Spikformer \cite{zhou2023spikformer}} & STBP Training & 4 & 95.19 & 4 & 77.86 & 16 & 80.6 \\
            & Transfer Learning & 4 & 97.03 & 4 & 83.83 & - & - \\
            SDT \cite{yao2023spike} & STBP Training & 4 & 95.6 & 4 & 78.4 & 16 & 80.0 \\
            SNN-ViT \cite{wang2025svit} & STBP Training & - & 96.1 & - & 80.1 & - & 82.3 \\
            Spikingformer \cite{zhou2026spikingformer} & STBP Training & 4 & 95.95 & 4 & 80.37 & 16 & 81.4 \\ \hline

            \multirow{3}{*}{\textbf{This Work}} & \textbf{Ge²mS-T Small} & 4 & \textbf{98.28} & 4 & \textbf{87.54} & 4 & \textbf{86.0} \\
            & \textbf{Ge²mS-T Base} & 4 & \textbf{98.45} & 4 & \textbf{88.73} & 4 & \textbf{87.5} \\
            & \textbf{Ge²mS-T Large} & 4 & \textbf{98.59} & 4 & \textbf{89.31} & 4 & \textbf{87.6} \\
            \Xhline{1pt}
    	\end{tabular}
    }
	\label{tab04}
\end{table*}
\begin{figure*}[h]
  \centering
  \begin{subfigure}[b]{0.32\linewidth}
    \centering
    \includegraphics[width=\linewidth]{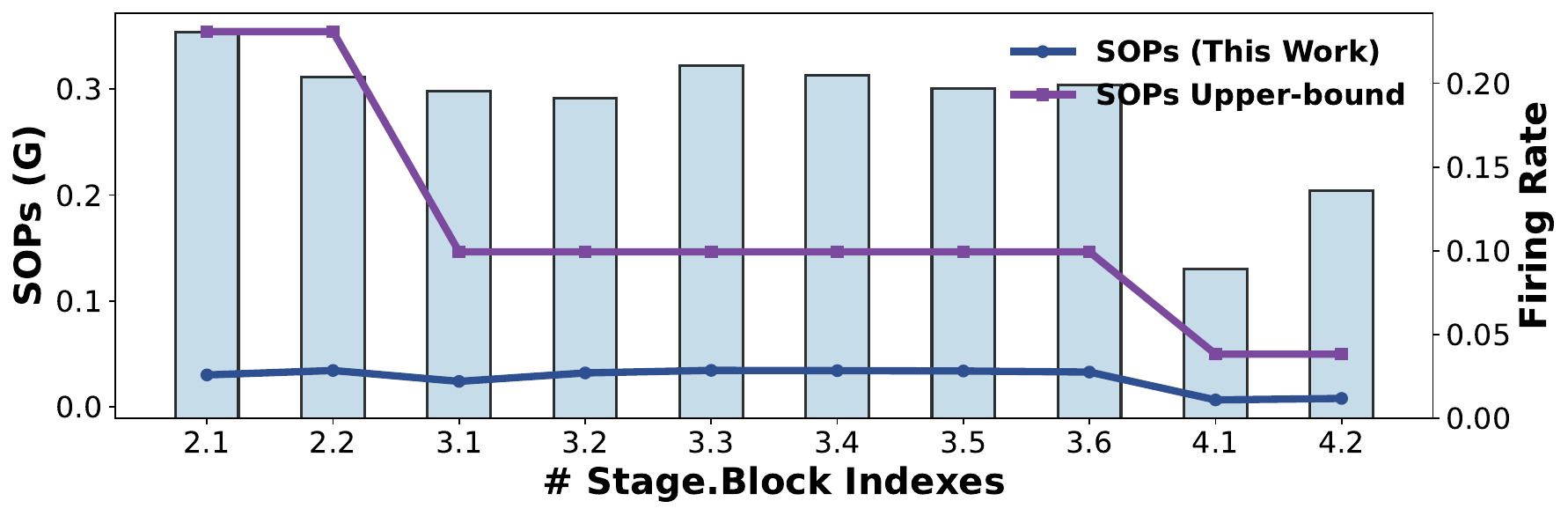}
    \caption{\footnotesize Ge²mS-T Small, SSA}
  \end{subfigure}
  \begin{subfigure}[b]{0.32\linewidth}
    \centering
    \includegraphics[width=\linewidth]{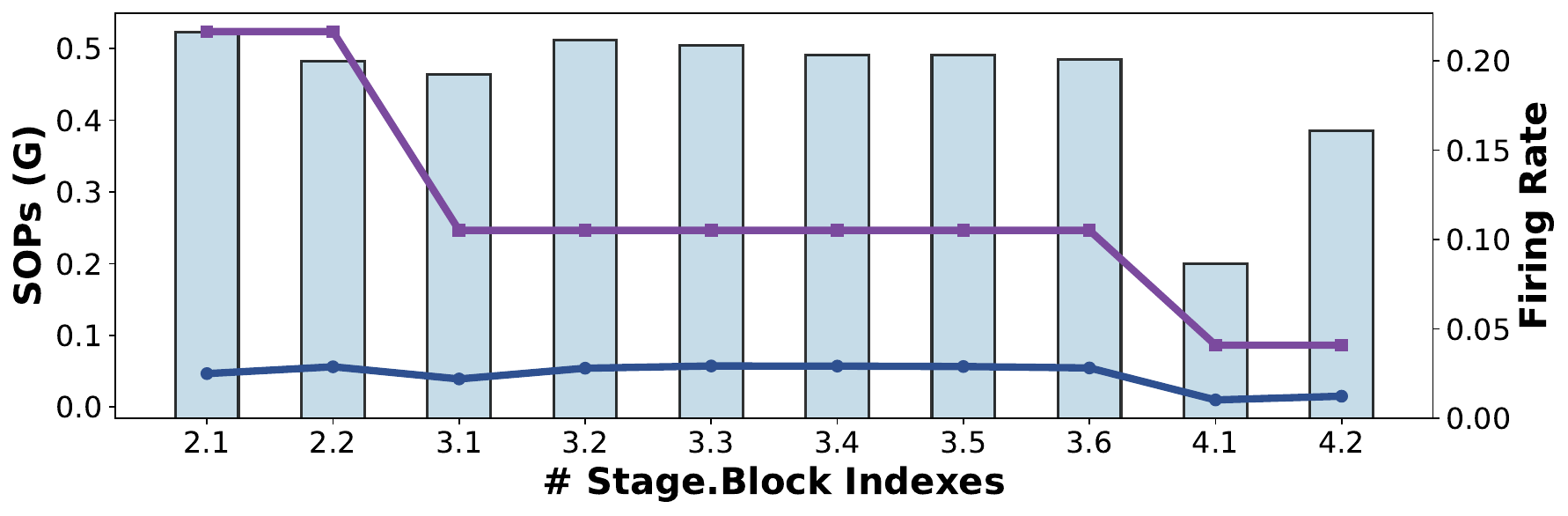}
    \caption{\footnotesize Ge²mS-T Base, SSA}
  \end{subfigure}
  \begin{subfigure}[b]{0.32\linewidth}
    \centering
    \includegraphics[width=\linewidth]{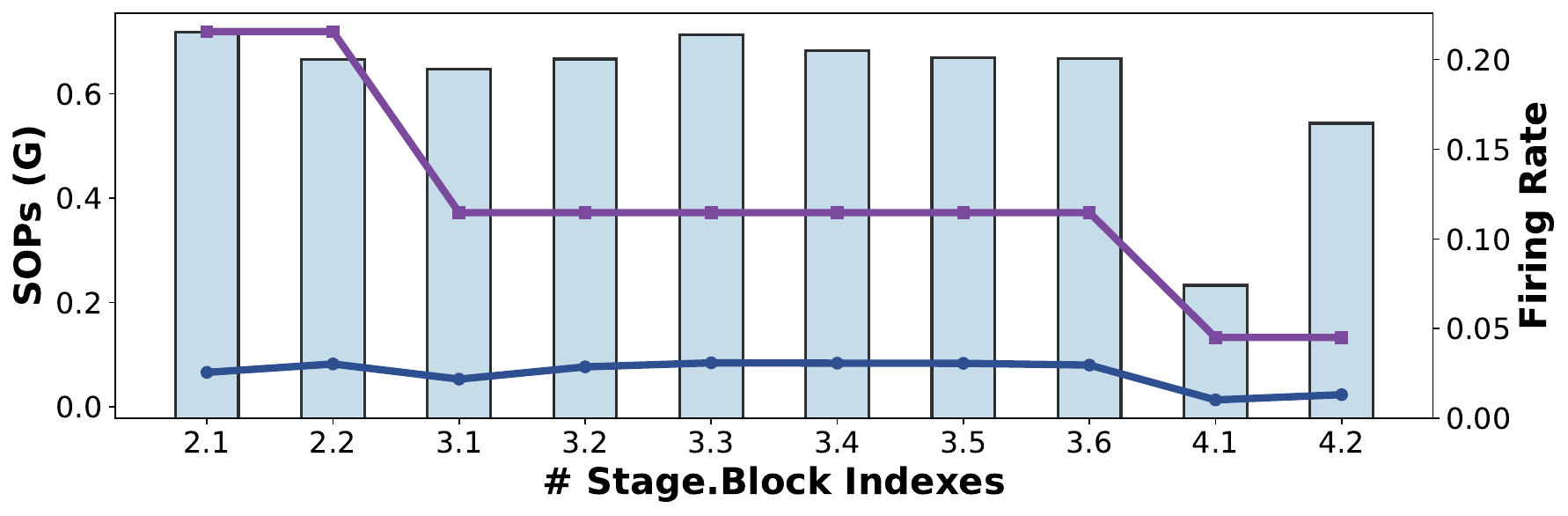}
    \caption{\footnotesize Ge²mS-T Large, SSA}
  \end{subfigure}
  \vspace{0.5em}
  \begin{subfigure}[b]{0.32\linewidth}
    \centering
    \includegraphics[width=\linewidth]{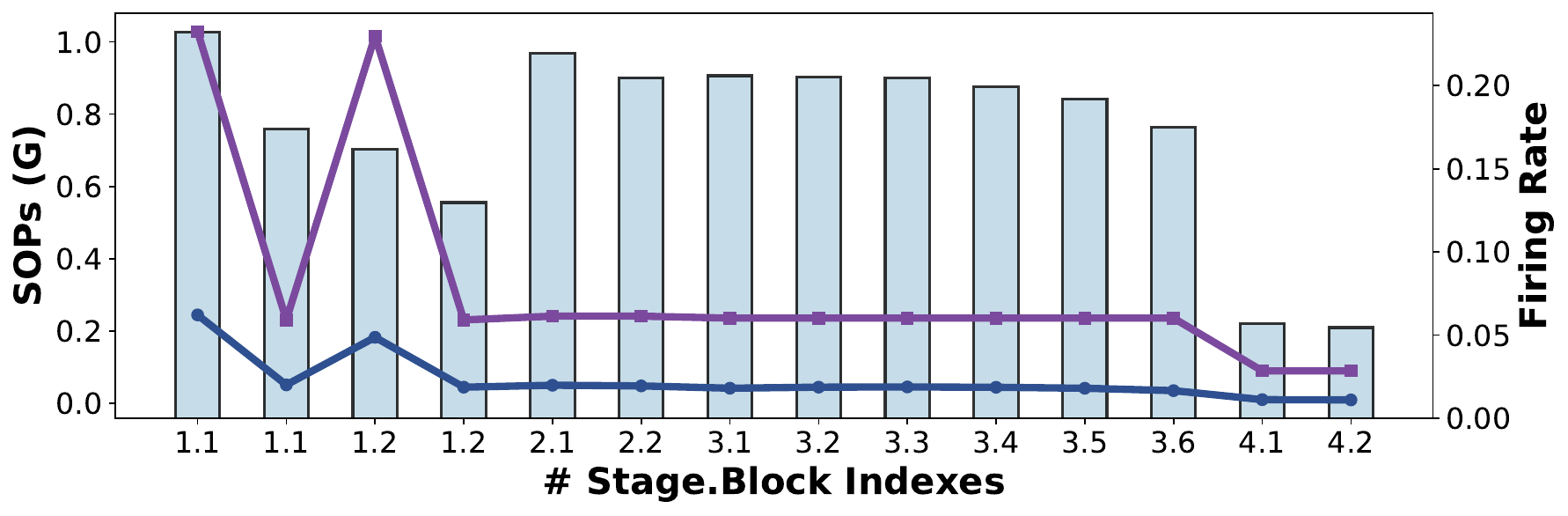}
    \caption{\footnotesize Ge²mS-T Small, SConv \& SFFN}
  \end{subfigure}
  \begin{subfigure}[b]{0.32\linewidth}
    \centering
    \includegraphics[width=\linewidth]{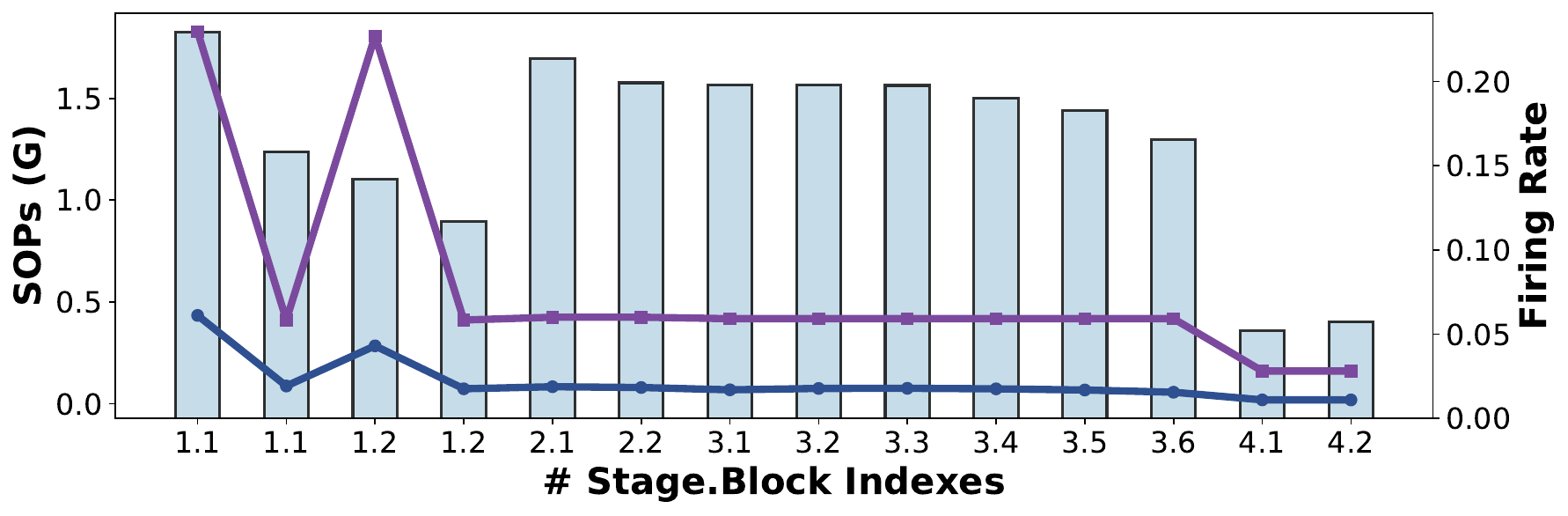}
    \caption{\footnotesize Ge²mS-T Base, SConv \& SFFN}
  \end{subfigure}
  \begin{subfigure}[b]{0.32\linewidth}
    \centering
    \includegraphics[width=\linewidth]{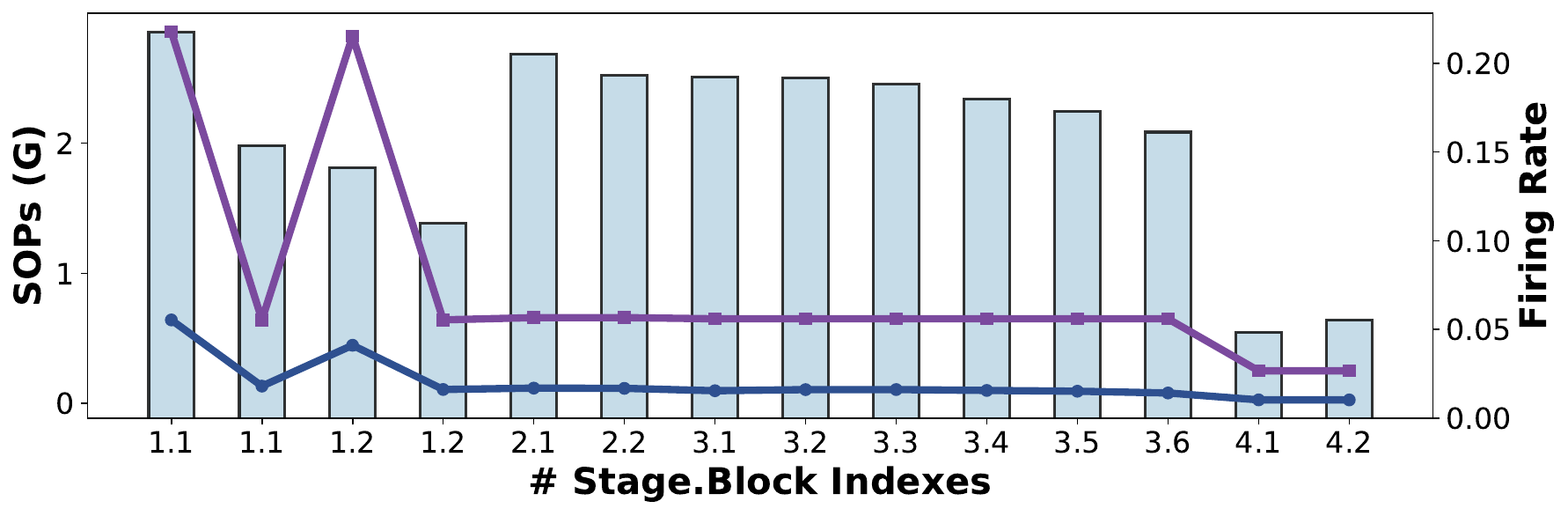}
    \caption{\footnotesize Ge²mS-T Large, SConv \& SFFN}
  \end{subfigure}
  \caption{Distribution of energy-related metrics for modules (SSA, SConv and SFFN) on the ImageNet-1k dataset, including average firing rate, SOPs and its theoretical upper-bounds.}
  \label{fig03}
\end{figure*}
Benefiting from the superior generalization capability of the pre-trained Ge²mS-T model trained on the large-scale ImageNet-1k dataset, we undertake further fine-tuning and subsequently validate the inference accuracy across a series of downstream datasets, as presented in Tab.\ref{tab04}.

\textbf{Static data-domain.} For CIFAR-10 and CIFAR-100 datasets, Ge²mS-T achieves performance superior to prior works employing diverse learning strategies and network architectures. For instance, Ge²mS-T Base surpasses the S-CNN-based methods TET \cite{deng2022temporal} and GAC-SNN \cite{qiu2024gated} by margins of 4.01\% and 2.21\% on CIFAR-10, respectively. Regarding CIFAR-100, the validated conclusions manifest even more significantly. On one hand, Ge²mS-T exceeds the Spikformer \cite{zhou2023spikformer} variants based on STBP Training and Transfer Learning by up to 11.45\% and 5.48\%, respectively. On the other hand, the inference accuracy of the Ge²mS-T family increases progressively with the growth of model parameters.

\textbf{Neuromorphic data-domain.} In contrast to static data, the representation of neuromorphic data is distributed across both temporal and channel dimensions. To effectively align with the input end of the pre-trained checkpoints, we incorporate an external convolutional layer to integrate the aforementioned temporal and channel dimensions and project them onto standard input dimensions before proceeding with fine-tuning. It is observed that Ge²mS-T requires only 4 inference time-steps to achieve accuracy improvements of up to 4.43\% and 6.2\% over TET \cite{deng2022temporal} and Spikingformer \cite{zhou2026spikingformer} on CIFAR10-DVS, respectively.

\subsection{Detailed Energy Consumption Statistics}
Figure \ref{fig03} elaborately illustrates the distribution of relevant energy consumption metrics in Ge²mS-T with modules serving as the fundamental computational units. Specifically, Fig.\ref{fig03}(a)-(c) depict the cumulative SOPs and the multi-layer averaged spiking firing rates across attention modules, wherein GW-SSA is employed in Stages 2-3 while SSA based exclusively on ExpG-IF is adopted in Stage 4. It can be observed that the magnitude of the SOPs Upper-bound in Stage 2 substantially exceed those in Stage 3, whereas the SOPs corresponding to our proposed approach remain consistently stable, suggesting that the adoption of a spatial-dimension-oriented grouping strategy ensures that the computational overhead of the attention mechanism is not significantly amplified with the increasing number of spiking tokens. With respect to Fig.\ref{fig03}(d)-(f), Stage 1 is constructed upon the two-layer SConv and SFFN modules as formulated in Eq.\eqref{eq07}, while Stages 4 utilizes SFFN based solely on ExpG-IF, with Conv-SFFN deployed in all remaining positions. It is noteworthy that compared to standard convolution operations, the depthwise separable convolution architecture of SFFN can substantially reduce SOPs, and the integration of implicit grouping oriented toward spike patterns can further extend this advantage.

\section{Conclusions}
In this work, we propose Ge²mS-T, a novel architecture grounded in triple-dimension grouping computation, which facilitates direct training with constant-order memory and enables precise regulation of spike firing patterns. It also exhibits advanced inference capabilities under conditions of extremely low computational overhead. We believe that the introduction of Ge²mS-T signifies a significant breakthrough in energy-efficient S-ViT architectures, which will further facilitate the deployment of SNNs on mobile devices and within resource-constrained environments.

\section*{Acknowledgments}
This work is supported by the National Natural Science Foundation of China (62422601,U24B20140, 62506011), Beijing Municipal Science and Technology Program (Z251100008125052), Beijing Nova Program (20240484703), China Postdoctoral Science Foundation (2026M791600), Qiyuan Innovative Talent Program, Beijing Key Laboratory of Brain-inspired Spiking Large Models, Peking University–Beijing Zhichuang JianYue Joint Laboratory for Embodied Intelligence.

\bibliographystyle{ACM-Reference-Format}
\bibliography{main}
\end{document}

\title{Ge²mS-T: Multi-Dimensional Grouping for Ultra-High Energy Efficiency in Spiking Transformer\\ -- Supplementary Materials --}

\author{Zecheng Hao}
\orcid{0000-0001-9074-2857}
\affiliation{
  \institution{Peking University}
  \department{School of Computer Science, State Key Laboratory for Multimedia Information Processing}
  \city{Beijing}
  \country{China}
}
\email{haozecheng@pku.edu.cn}

\author{Shenghao Xie}
\orcid{0009-0003-1024-1166}
\affiliation{
  \institution{Peking University}
  \department{School of Computer Science, State Key Laboratory for Multimedia Information Processing}
  \city{Beijing}
  \country{China}
}
\email{shenghaoxie@stu.pku.edu.cn}

\author{Kang Chen}
\orcid{0009-0001-2161-0364}
\affiliation{
  \institution{Peking University}
  \department{School of Computer Science, State Key Laboratory for Multimedia Information Processing}
  \city{Beijing}
  \country{China}
}
\email{mrchenkang@stu.pku.edu.cn}

\author{Wenxuan Liu}
\orcid{0000-0002-4417-6628}
\authornote{Corresponding author.}
\affiliation{
  \institution{Peking University}
  \department{School of Computer Science, State Key Laboratory for Multimedia Information Processing}
  \city{Beijing}
  \country{China}
}
\email{lwxfight@126.com}

\author{Zhaofei Yu}
\orcid{0000-0003-0683-6936}
\affiliation{
  \institution{Peking University}
  \department{School of Computer Science, Beijing Key Laboratory of Brain-inspired Spiking Large Models}
  \city{Beijing}
  \country{China}
}
\email{yuzf12@pku.edu.cn}

\author{Tiejun Huang}
\orcid{0000-0002-4234-6099}
\affiliation{
  \institution{Peking University}
  \department{School of Computer Science, State Key Laboratory for Multimedia Information Processing}
  \city{Beijing}
  \country{China}
}
\email{tjhuang@pku.edu.cn}

\renewcommand{\shortauthors}{Zecheng Hao et al.}



\maketitle
\appendix
\setcounter{equation}{0}
\setcounter{figure}{0}
\setcounter{table}{0}
\renewcommand{\thetable}{S\arabic{table}}
\renewcommand{\thefigure}{S\arabic{figure}}
\renewcommand{\theequation}{S\arabic{equation}}

\section{Efficiency-Aware Performance Analysis}
\begin{table}[h]
    \caption{Efficiency-aware performance comparison using accuracy-to-parameter (A/P) and accuracy-to-energy (A/E) ratios on the ImageNet-1k dataset.}
    \renewcommand\arraystretch{1.0}
    \centering
    \resizebox{\linewidth}{!}{
    \begin{threeparttable}
    \begin{tabular}{cc|cc} \Xhline{1pt}
        \textbf{Method} & \textbf{Architecture} & \textbf{A/P (\%/M)} & \textbf{A/E (\%/mJ)} \\ \hline
        Spiking ResNet & ResNet-34 & 3.29 & 1.21 \\
        Spiking ResNet & ResNet-50 & 2.85 & 1.03 \\
        STBP-tdBN & Spiking ResNet-34 & 2.92 & 9.97 \\
        SEW ResNet & SEW ResNet-34 & 3.08 & 16.59 \\
        SEW ResNet & SEW ResNet-50 & 2.65 & 13.86 \\
        SEW ResNet & SEW ResNet-101 & 1.54 & 7.72 \\
        SEW ResNet & SEW ResNet-152 & 1.15 & 5.37 \\
        MS ResNet & MS ResNet-34 & 3.19 & 16.18 \\
        MS ResNet & MS ResNet-104$^\dag$ & 0.98 & 7.46 \\
        Att-MS ResNet & MS ResNet-104$^\dag$ & 0.98 & 10.56 \\ \hline
        Spikformer & Spikformer-8-384 & 4.18 & 9.09 \\
        Spikformer & Spikformer-8-512 & 2.47 & 6.34 \\
        Spikformer & Spikformer-8-768 & 1.13 & 3.48 \\
        Spike-driven Transformer & SDT-8-384 & 4.30 & 18.53 \\
        Spike-driven Transformer & SDT-8-512 & 2.51 & 16.57 \\
        Spike-driven Transformer & SDT-8-768 & 1.15 & 12.53 \\
        SNN-ViT & SViT-8-256 & 5.45 & 5.23 \\
        SNN-ViT & SViT-8-384 & 2.53 & 3.69 \\
        Spikingformer & Spikingformer-8-384 & 4.42 & 13.25 \\
        Spikingformer & Spikingformer-8-512 & 2.58 & 8.82 \\
        Spikingformer & Spikingformer-8-768 & 1.17 & 4.76 \\ \hline
        \textbf{Ge²mS-T} & \textbf{Ge²mS-T Small} & \textbf{14.04} & \textbf{64.76} \\
        \textbf{Ge²mS-T} & \textbf{Ge²mS-T Base} & \textbf{8.35} & \textbf{40.72} \\
        \textbf{Ge²mS-T} & \textbf{Ge²mS-T Large} & \textbf{5.51} & \textbf{28.20} \\
        \Xhline{1pt}
    \end{tabular}
    \begin{tablenotes}
        \small
        \item[$\dag$] denotes enlarged 288$\times$288 input size for inference; default is 224$\times$224 for training and inference stages.
    \end{tablenotes}
    \end{threeparttable}
    }
    \label{tabS1}
\end{table}
\begin{table*}[t]
    \caption{Experimental configuration of Ge²mS-T for data processing and training on various benchmarks.}
    \renewcommand\arraystretch{1.08}
	\centering
    \resizebox{\linewidth}{!}{
    	\begin{tabular}{c|ccc|ccc} \Xhline{1pt}
            \multirow{2}{*}{\textbf{Experimental Configuration}} & \multicolumn{3}{c|}{\textbf{ImageNet-1k}} & \multirow{2}{*}{\textbf{CIFAR-10}} & \multirow{2}{*}{\textbf{CIFAR-100}} & \multirow{2}{*}{\textbf{CIFAR10-DVS}} \\ \cline{2-4}
            & \textbf{Ge²mS-T Small} & \textbf{Ge²mS-T Base} & \textbf{Ge²mS-T Large} & & & \\ \hline
            Batch Size & \multicolumn{3}{c|}{8 $\times$ 64} & \multicolumn{2}{c}{64} & 16 \\
            \multirow{3}{*}{Data Augmentation} & \multicolumn{3}{c|}{Random Crop, Horizontal Flip, Color Jitter} & \multicolumn{3}{c}{Horizontal Flip} \\
            & Rand-m7-mstd0.5-inc1 & \multicolumn{2}{c|}{Rand-m9-mstd0.5-inc1} & \multicolumn{2}{c}{Rand-m9-n1-mstd0.4-inc1} & Rand-(Roll, Rotation, Cutout)-n1 \\
            & \multicolumn{3}{c|}{Mixup \& CutMix} & \multicolumn{3}{c}{Mixup \& CutMix} \\ \hline
            Optimizer & \multicolumn{3}{c|}{AdamW, $lr=5\times10^{-4}, wd=10^{-2}$} & \multicolumn{3}{c}{AdamW, $lr=10^{-4}, wd=10^{-2}$} \\
            Scheduler & \multicolumn{3}{c|}{Cosine Annealing, $lr_{min}=10^{-5}$} & \multicolumn{3}{c}{Cosine Annealing, $lr_{min}=10^{-5}$} \\
            Training Epochs & \multicolumn{3}{c|}{300} & \multicolumn{3}{c}{100} \\
            \Xhline{1pt}
    	\end{tabular}
    }
	\label{tabS3}
\end{table*}
As summarized in Tab.\ref{tabS1}, the proposed Ge²mS-T framework demonstrates consistent superiority in resource-normalized efficiency across both S-CNN and S-ViT families. Within the S-CNN category, Ge²mS-T Small establishes a new efficiency frontier with an accuracy-to-parameter ratio of 14.04 \%/M and accuracy-to-energy ratio of 64.76 \%/mJ—values that exceed those of Spiking ResNet-34 by factors of 4.27 and 53.52, respectively. When benchmarked against the most parameter-heavy SEW ResNet-152, Ge²mS-T Base delivers 7.26$\times$ and 7.58$\times$ gains in the two normalized metrics while simultaneously achieving a 12.89\% relative accuracy uplift. Furthermore, Ge²mS-T Large not only surpasses the attention-enhanced MS ResNet-104 by 3.80\% in raw accuracy, but also attains 5.62$\times$ higher accuracy per parameter and 2.67$\times$ higher accuracy per unit energy relative to its attention-augmented counterpart.

Turning to S-ViT architectures, Ge²mS-T Small redefines the efficiency-accuracy trade-off: its normalized scores (14.04 \%/M for A/P, 64.76 \%/mJ for A/E) eclipse those of the largest Spikformer variant by more than an order of magnitude, despite operating with substantially fewer parameters and lower energy budget. Against the spike-driven Transformer baseline (SDT-8-384), Ge²mS-T Base nearly doubles the accuracy-per-parameter metric (1.94$\times$) and improves accuracy-per-energy by 120\%, all while maintaining a 5.91\% absolute accuracy advantage. Even when compared to the compact SViT-8-256, Ge²mS-T Base achieves a 53\% gain in parameter efficiency and a 7.79$\times$ improvement in energy efficiency. Most notably, Ge²mS-T Large secures the highest absolute accuracy (79.82\%) among all evaluated models while sustaining accuracy-to-parameter and accuracy-to-energy ratios that are 4.71$\times$ and 5.92$\times$ those of Spikingformer-8-768, underscoring the effectiveness of the Ge²mS-T training paradigm in jointly optimizing learning performance and computational resource utilization across model scales.

\section{Ablation Studies for the Grouping Strategies of Ge²mS-T Framework}
\begin{table}[h]
    \caption{Performance comparison of Ge²mS-T Large, CIFAR-100 under various grouping strategies.}
    \renewcommand\arraystretch{1.0}
    \centering
    \resizebox{\linewidth}{!}{
    \begin{tabular}{cc|cc} \Xhline{1pt}
        \multicolumn{2}{c|}{\textbf{Component Configuration}} & \multirow{2}{*}{\textbf{Energy(mJ)}} & \multirow{2}{*}{\textbf{Acc.(\%)}} \\ \cline{1-2}
        \textbf{Temporal G.} & \textbf{Spatial \& Structural G.} & & \\ \hline
        \textcolor{green}{\ding{51}} & \textcolor{green}{\ding{51}} & 2.87 & 89.31 \\ \hline
        \textcolor{green}{\ding{51}} & \textcolor{red}{\ding{55}} & 3.02 & 87.80 \\
        \textcolor{red}{\ding{55}} & \textcolor{green}{\ding{51}} & 3.80 & 89.30 \\
        \textcolor{red}{\ding{55}} & \textcolor{red}{\ding{55}} & 4.00 & 87.90 \\
        \Xhline{1pt}
    \end{tabular}
    }
    \label{tabS2}
\end{table}
In Tab.\ref{tabS2}, we conduct experimental validation on the downstream dataset CIFAR-100 based on the pre-trained backbone from Ge²mS-T-Large, ImageNet-1k, further exploring the performance changes of our framework under the condition of merely selecting partial grouping strategies. It can be observed that after disabling the grouping strategy for spike sequences, the accuracy of Ge²mS-T does not achieve significant improvement, but the corresponding energy consumption increases significantly by 0.93-0.98 mJ. While after disabling GW-SSA, the energy consumption and accuracy of Ge²mS-T increases by 0.15-0.20 mJ and decreases by 1.40\%-1.51\%, respectively. The above results indicate that simultaneously enabling multi-dimensional grouping strategies does not have a negative impact for this framework. On the contrary, it produces a synergistic effect, achieving further performance advantages.

\section{Detailed Experimental Configuration}
\textbf{Hardware and Batch Size.}
As shown in Tab.\ref{tabS3}, training procedure on the ImageNet-1k dataset is conducted on 8 NVIDIA H100 80GB GPUs with a batch size of 64 per GPU. For all other datasets, experiments are performed on a single NVIDIA GPU, using a batch size of 64 for CIFAR-10(100) and 16 for CIFAR10-DVS.

\textbf{Data Augmentation.}
We adopt tailored augmentation strategies for static and neuromorphic benchmarks:
\begin{itemize}
    \item \textbf{ImageNet-1k:} The pipeline includes Random Crop, Horizontal Flip and Color Jitter. Additionally, RandAugment~\cite{cubuk2020randaugment} is applied with magnitude scheduling enabled. Specifically, Ge²mS-T Small uses a magnitude level of 7 with a standard deviation of 0.5, while the Ge²mS-T Base and Large variants adopt a magnitude level of 9 with a standard deviation of 0.5.
    \item \textbf{CIFAR-10(100):} We apply Horizontal Flip combined with RandAugment, configured with a magnitude level of 9, one operation per image, a standard deviation of 0.4 and magnitude scheduling.
    \item \textbf{CIFAR10-DVS:} Due to the sparse nature of event data, we employ a custom geometric strategy where one transformation is randomly selected from Roll (offset ratio $\approx$ 0.33), Rotation (degree $\leq 30^\circ$) or Cutout~\cite{devries2017improved} (area ratio $\approx$ 0.33), complemented by Random Horizontal Flip.
    \item \textbf{Mixup \& CutMix:} For all training data, we jointly apply Mixup~\cite{zhang2017mixup} ($\alpha=0.8$) and CutMix~\cite{yun2019cutmix} ($\alpha=1.0$) with a switching probability of 0.5 during batch construction.
\end{itemize}

\textbf{Optimizer and Scheduler.}
All models are optimized using the AdamW optimizer~\cite{loshchilov2017decoupled} with a weight decay of $10^{-2}$. The initial learning rate is set to $5\times10^{-4}$ for ImageNet-1k and $10^{-4}$ for the remaining datasets. A Cosine Annealing scheduler~\cite{loshchilov2016sgdr} is employed across all experiments, decaying the learning rate to a minimum of $10^{-5}$ over 300 epochs for ImageNet-1k and 100 epochs for the other datasets.

\bibliographystyle{ACM-Reference-Format}
\bibliography{main}